\documentclass[a4paper]{article}

\usepackage[T1]{fontenc}
\usepackage[utf8]{inputenc}
\usepackage{combelow}

\usepackage{graphicx}
\usepackage{url}
\usepackage{natbib}
\usepackage{xcolor}

\title{RELATE: A Modern Processing Platform for Romanian Language}

\author{
Vasile P\u{a}i\cb{s} \and
Radu Ion \and
Andrei-Marius Avram \and
Maria Mitrofan \and
Dan Tufi\cb{s}
}

\date{
Research Institute for Artificial Intelligence "Mihai Dr\u{a}g\u{a}nescu", Romanian Academy \\
              Tel.: +40-21-3188106\\
              Fax: +40-21-3188142\\
              \{vasile,radu,andrei.avram,maria,tufis\}@racai.ro
              }

\begin{document}

\maketitle

\begin{abstract}
This paper presents the design and evolution of the RELATE\footnote{\url{https://relate.racai.ro}} platform. It provides a high-performance environment for natural language processing activities, specially constructed for Romanian language. Initially developed for text processing, it has been recently updated to integrate audio processing tools. Technical details are provided with regard to core components. We further present different usage scenarios, derived from actual use in national and international research projects, thus demonstrating that RELATE is a mature, modern, state-of-the-art platform for processing Romanian language corpora. Finally, we present very recent developments including bimodal (text and audio) features available within the platform.
\end{abstract}

\section{Introduction}
\label{sec:intro}
Language Technology (LT) platforms provide services for analysis or production of written or spoken language. They employ artificial intelligence (AI) methods to implement their functionalities. Furthermore, platforms have been developed either to showcase new technologies or as powerful tools used in processing large amounts of data, in the form of offline corpora or online requests.

One of the motivations of the 1st International Workshop for Language Technology Platforms (IWLTP 2020) was to address the issue of fragmentation in the Language Technology landscape. As the organizers note \citep{iwltp-2020-international}, instead of competing with one another, platforms should be constructed to be interoperable and interact with each other to create synergies towards a productive LT ecosystem. Interoperability is usually achieved by providing input and output in standardized formats, specific to the tasks being executed. Interaction between systems involves exposing one system's functionality in a way that can be used from another system.

In this paper we present the RELATE platform \citep{pais2020relatelrec}. It is a modular, modern, state-of-the-art platform for processing Romanian language, developed at the Institute for Artificial Intelligence "Mihai Dr{\u{a}}g{\u{a}}nescu" of the Romanian Academy. It integrates both technologies developed in our institute and technologies developed by partner institutions. RELATE is being used actively in multiple national and international research projects. From the beginning it was designed to use standardized file formats, thus ensuring interoperability with other language processing systems. Internal functions are available as REST web services, thus allowing for interaction with other systems. 

This paper is organized as follows: Section \ref{sec:related} presents related work, Section \ref{sec:history} presents the history behind RELATE platform development, then Section \ref{sec:architecture} introduces its architecture, while Section \ref{sec:components} describes the available components. Section \ref{sec:scenarios} presents several usage scenarios. Finally we conclude in Section \ref{sec:conclusions}.

\section{Related work}
\label{sec:related}

META-SHARE\footnote{\url{http://www.meta-share.org}} \citep{federmann-etal-2012-meta}, CLARIN\footnote{\url{https://www.clarin.eu}} and ELRC-Share\footnote{\url{https://elrc-share.eu/}} (European Language Resource Coordination Share) are publicly available European websites for research and development in the field, allowing access to language resources. Both ELRC-Share and META-SHARE offer advanced search facilities through which one can easily find various language tools and corpora (text, annotated, audio corpora, etc.) for any (European and other) language. Complex processing pipelines, like NLP-Cube \citep{boros-etal-2018-nlp} and TTL are able to perform tokenization, lemmatization, POS tagging, chunking and dependency parsing, but require programming knowledge in order to integrate them into an application. More recently, the TEPROLIN web service integrates multiple tools, including NLP-Cube, TTL and solutions for named entity recognition \citep{pais2019phd} and biomedical named entity recognition, into an easy to use pipeline. However usage of this web service still requires programming knowledge for web service communication.

GATE \citep{cunningham-etal-2002-gate} and TextFlows \citep{PEROVSEK2016128} aim to make the composition of the language processing chains more user-friendly. For this purpose, the graphical interfaces allow for dragging and dropping text processing widgets into a graphical processing workflow. However, their output is not enhanced with specialized visualization tools that allow access into the computational resources used for annotation. Moreover, they only focus on purely textual content, while multimodal content (such as text and audio) is not handled within the platforms.

CoBiLiRo \citep{cristea-etal-2020-cobiliro} is a storage platform for multimodal (text and audio) corpora. It was developed in the context of the RETEROM\footnote{\url{https://racai.ro/p/reterom/}} project for storing Romanian speech corpora suitable for the project's purposes. It allows defining a large number of metadata fields, but it doesn't allow for any complex language processing tasks.

RELATE aims specifically at doing automatic text processing, with annotations at multiple levels, along with annotation visualization and expansion into the corresponding linguistic computational resources. The platform focuses on interactive user experience, via web based interfaces. Compared to other platforms, such as \citep{che-etal-2010-ltp}, the main focus of RELATE is not on exposing APIs, even though API access to platform functionality is available. Text processing APIs are used internally, accessible by platform components, and most of them can also be invoked externally.

RELATE makes use of a common internal format, comprising multiple files (text, standoff metadata, CoNLL-U Plus\footnote{\url{https://universaldependencies.org/ext-format.html}} annotations). The processing workflow is constructed by adding individual tasks. Each task is able to both use and produce data according to the internal format. For this reason no workflow editor, such as the one used in \citep{PEROVSEK2016128}, is currently available.

\citep{coleman-etal-2020-architecture} presents a platform for integrating multiple Machine Translation (MT) models. In RELATE we only provide MT capabilities related to the Romanian language (currently allowing only Romanian-English and English-Romanian translation). We further allow translated content to become an input for further language processing tasks. \citep{rebai-etal-2020-linto} presents a platform integrating a voice assistant for improving efficiency and productivity in business. In the case of RELATE, we only integrate existing Romanian (and English) Automatic Speech Recognition (ASR) and Text-to-Speech (TTS) systems while still focusing on the textual component. Thus ASR output can be passed as input to further language processing tasks. 

\citep{rehm-etal-2020-towards} acknowledges that a large number of AI platforms are currently under development, both on the national level, supported through local funding programmes, and on the international level, supported by the European Union. The authors further recognize the enormous fragmentation of the European AI and LT landscape and consider that modern platforms should be able to exchange information, data and services, in order to enable interoperability. We agree with this assessment and thus in the RELATE platform we aimed to make use of standardized formats as well as decoupling functionality into components, that could be invoked externally if needed. In addition, similar to the AI4EU\footnote{\url{https://www.ai4eu.eu/}} and European Language Grid (ELG)\footnote{\url{https://www.european-language-grid.eu/}} platforms, RELATE is able to integrate Docker containers for language processing tools. However, this is not imposed and thus only a small number of the available components are integrated as containers.

\section{Evolution of the RELATE platform}
\label{sec:history}

The RELATE platform was developed across a number of national and international research projects and evolved together with the requirements associated with the activities it was used for. This section describes its evolution throughout the projects, from the first implementation to the current version.

RELATE platform development \citep{pais2019relate} started within the RE\-TE\-ROM project. This was a national project, started in 2018. One of the primary goals, associated with the sub-project TEPROLIN, was to develop and integrate state of the art technologies for Romanian natural language processing, such as tokenization, part-of-speech tagging, dependency parsing, phonetic annotations, named entity recognition. The primary result was the TEPROLIN web service \citep{ion2018teprolin} which allows invocation using a raw text document and produces different levels of annotations (based on specified parameters) encoded using a custom JSON format\footnote{\url{http://relate.racai.ro/?path=teprolin/doc_dev}}.

The first implementation of the RELATE platform allowed for the management of large corpora  (upload, download, editing) and parallel processing using the TEPROLIN service. For processing purposes a task management engine was implemented, allowing distribution of documents across any number of TEPROLIN processes started on the same server or over the network. This allowed for increased processing speed when needed, as the number of processing nodes can be adjusted dynamically. Furthermore, the platform is able to convert between the custom JSON encoding associated with TEPROLIN to a more standard CoNLL-U format\footnote{\url{https://universaldependencies.org/format.html}}, used also by the Universal Dependencies project\footnote{\url{https://universaldependencies.org/}}, or its extension CoNLL-U Plus\footnote{\url{https://universaldependencies.org/ext-format.html}} when appropriate.

Graphical representations were developed to allow a human user to manually explore annotation results. These include: visualization in different column-based and JSON formats, tree-based representations for dependency parsing, highlighting of recognized named entities. In order to improve the user's experience, we also included different interface elements allowing to query the Representative Corpus of Contemporary Romanian Language (CoRoLa) \citep{tufis2019corola}. This includes links to the main query interface of the text component of CoRoLa, using the KorAP corpus analysis platform \citep{banski-etal-2012-new}, and to the speech component, allowing searching in audio files \citep{boros2018speechtools} and listening to words being pronounced by Romanian speakers. Different visualizations are also making use of word embeddings \citep{bojanowski-etal-2017-enriching} computed on the CoRoLa corpus \citep{pais2018corolawe} to suggest words appearing in similar contexts.

The Romanian WordNet \citep{Tufis2015wordnet} was integrated in order to allow words to be searched online. We further exploited the alignment between the Romanian and English \citep{miller1995wordnet} wordnets in order to provide aligned queries based on synset identifiers. 


The textual translation component, developed by TILDE with the involvement of the Research Institute for Artificial Intelligence "Mihai Drăgănescu" of the Romanian Academy, within the project "CEF Automated Translation toolkit for the Rotating Presidency of the Council of the EU", was integrated in RELATE, by means of the TILDE Machine Translation API\footnote{\url{https://www.tilde.com/developers/machine-translation-api}}. This allows users to translate documents directly in the platform. After a successful translation from English to Romanian, the resulting document can be analysed using the platform's functionality.

Within the "Multilingual Resources for CEF.AT in the legal domain" (MARCELL) project\footnote{\url{https://marcell-project.eu/}}, a terminology annotation tool \citep{coman2019} was developed and integrated in RELATE allowing for identification of terms from EuroVoc\footnote{\url{https://eur-lex.europa.eu/browse/eurovoc.html}} and IATE\footnote{\url{https://iate.europa.eu/home}} (Interactive Terminology for Europe) terminologies. The tool can be used following a lemmatization and part-of-speech tagging operation.

The ASR system \citep{avram2020towards} resulted from the national project ROBIN\footnote{\url{http://aimas.cs.pub.ro/robin/en/}}, initially developed for human-robot interaction \citep{ion2020dialoguemanager,tufis2019pepper},  was integrated in the RELATE platform. This allows text to be extracted either from directly recorded speech or from uploaded audio files. The resulting text can then be analyzed using the platform's processing tools. Several corrections on the recognized text (such as basic comma restoration and truecasing) can be performed \citep{avram2020robinspeech} before the actual analysis. 

We further exploited the availability of ASR and text translation features in order to provide speech-to-speech translation functionality for Romanian-English and English-Romanian. This chained together different platform modules and required the integration of additional text-to-speech components. To synthesize the Romanian speech, we integrated two models in our pipeline: Romanian TTS developed by \cite{stan2011romanian} and RACAI SSLA developed by \cite{boros2018speechtools}, that are based on Hidden Markov Models (HMM) to compute the most probable sequence of spectrograms.

The project "Curated Multilingual Language Resources for CEF.AT" (CURLICAT), started in 2021, requires large scale anonymization of the collected text corpus. This process was further integrated in RELATE, making use of the parallelization mechanism already available for different text processing tasks. 

One of the research directions in our institute, in line with the COST Action "Nexus Linguarum", aims to create resources specific to Linguistic Linked Open Data specifications \citep{mititelu2020llod}. This provided an opportunity for development and inclusion in RELATE  of post-processing capabilities allowing RDF exports for different annotation levels. 

The "European Language Equality" (ELE) project\footnote{\url{http://www.european-language-equality.eu}}, started in 2021, has as primary goals the development of a sustainable evidence-based strategic research agenda and roadmap setting out actions, processes, tools, and actors to achieve full digital language equality of all languages (official or otherwise) used within the European Union through the effective use of language technologies. In addition, the research agenda encompasses the determination of the current state of language technologies and language equality within the EU. In turn, this led to an extension of  RELATE allowing multiple similar text processing pipelines \citep{pais2020multiplepipelinesrelate} to be executed on the same corpus in order to compare the results (with or without the presence of a "gold" annotated corpus).

\section{Platform architecture}
\label{sec:architecture}
The RELATE platform is organized following a modular approach. Each component is implemented independent of other components and provides a JSON descriptor exposing the corresponding menu entries and functionality at different user access levels. Moreover, components are service oriented. They can consume web services (either SOAP or REST) and expose additional HTTP REST services. In turn, the exposed services can be consumed by the graphical user interface of the component via AJAX calls to the REST endpoints. 

The service oriented architecture being employed allows for easy reuse of component provided functionality in other parts of the platform or even from outside the platform. By consuming services exposed by processing components it is possible to deploy these functions on different computing nodes, thus enabling high-speed parallel processing across interconnected servers. This scenario was employed for large corpora annotation in our institute, employing multiple servers available in the same local area network (LAN). However, since a processing operation takes more time than communication, it is possible to use remote hardware resources (across the Internet) for further increasing the processing capabilities available to the platform.

\begin{figure*}
\includegraphics[width=0.75\textwidth]{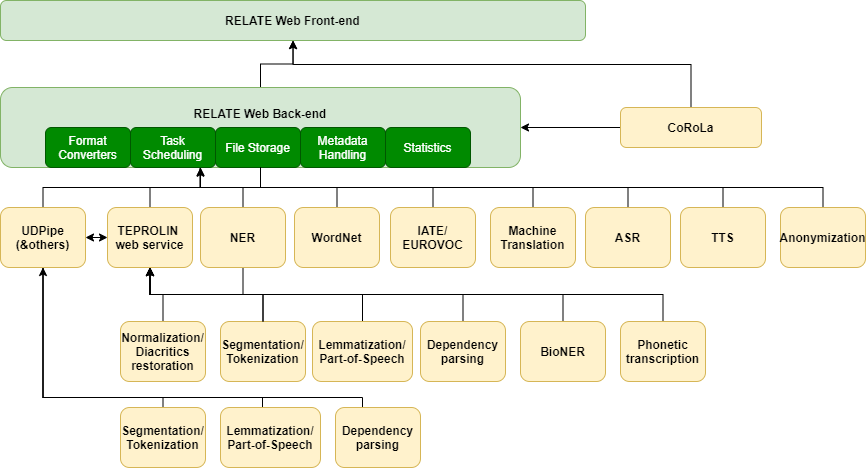}
\caption{RELATE platform architecture}
\label{fig:architecture}
\end{figure*}

Figure \ref{fig:architecture} presents the general platform architecture with the different components. The front-end part of the platform is in charge of user interaction. It provides visualizations for the different components. It is implemented using modern front-end technologies and languages such as HTML5, CSS, JavaScript. Communication with the platform's back-end is realized via page loading or AJAX calls when possible. The use of AJAX enhances the user experience by reducing loading times. Furthermore, exposure of different APIs as HTTP REST services allows for direct usage from outside the platform if needed.

The RELATE back-end is in charge of communication with different language-specific components. The back-end itself is written in PHP, while the components can be written in any language as long as a communication mechanism is possible. The platform currently has components written in C++, Java and Python. The preferred communication mechanism is via HTTP REST API calls. Nevertheless, the platform integrated also components called by invoking new processes with command line arguments.

\section{Primary platform components}
\label{sec:components}

This section will present the primary platform components (as indicated in Figure \ref{fig:architecture}), covering topics such as functions provided and implementation details. Only components with recent developments are included. Older components, such as the WordNet and CoRoLa query interfaces are not covered in detail. Additionally, some of the smaller (or very specific) components will be covered in Section \ref{sec:scenarios}.

\subsection{Corpus annotation}
\label{sec:annotation}

TEPROLIN \citep{ion2018teprolin} is a text pre-processing platform, implemented as a web service. It was developed in the ReTeRom project\footnote{\url{https://www.racai.ro/p/reterom/}} with the specific intent of making it easy to incorporate diverse text pre-processing tools written in a host of programming languages. TEPROLIN is written in Python 3 and requires each participating text pre-processing application to implement the TEPROLIN API class\footnote{\url{https://github.com/racai-ai/TEPROLIN/blob/master/TeproApi.py}}. The following are required for best results in TEPROLIN:
\begin{itemize}
    \item the loading of external resources is always made when the object is created so that with large resource files, the loading time does not interfere with the run time;
    \item text pre-processing applications should be resident processes on the machine running TEPROLIN and the I/O with TEPROLIN should be made with available IPC mechanisms, such as TCP sockets;
    \item if text pre-processing applications are implemented as web services, no resource loading is necessary and the \texttt{\_runApp} method is the only one which needs implementing.
\end{itemize}

TEPROLIN can be used as a standalone Python 3 object or from  RELATE, in demo\footnote{\url{https://relate.racai.ro/index.php?path=teprolin/custom}} or production mode (for annotating large corpora). Each processing operation (e.g. POS tagging, dependency parsing, etc.) can be configured to be executed with the specified TEPROLIN module able to perform it and, if desired, only a set of operations can be requested to be performed. TEPROLIN will auto-configure to execute all other required operations, such that the requested ones can be performed. For instance, if the user requires POS tagging, TEPROLIN will automatically execute sentence splitting and tokenization beforehand.

Currently, TEPROLIN incorporates the following text pre-processing applications:
\begin{itemize}
    \item NLP-Cube \citep{boros-etal-2018-nlp}, UDPipe \citep{udpipe2016straka} and TTL \citep{ion2007phd} for sentence splitting, POS tagging and lemmatization;
    \item additionally, NLP-Cube and UDPipe perform dependency parsing while TTL performs noun phrase, verb phrase, prepositional phrase, adverbial and adjectival phrase non-recursive chunking;
    \item MLPLA \citep{boros2018speechtools} for word hyphenation, stressed accented syllable detection and phonetic transcription;
    \item in house developed tools for Romanian text normalization, diacritic restoration, abbreviation and numeral expansion (i.e. automatically derive the full form of the abbreviation or the literal transcription of the numeral);
    \item BioNER \citep{mitrofan2018biomedical} based on a previous version of NLP-Cube adapted for NER processing;
    \item NER \citep{pais2019phd} for detecting names of persons, organizations, locations and time expressions.
\end{itemize}

UDPipe\footnote{\url{https://github.com/ufal/udpipe}} \citep{udpipe2016straka} is a trainable application for tokenization, tagging, lemmatization and dependency parsing of CoNLL-U files. Sentence splitting and tokenization are performed by checking each character in the input if it's the end of a token or the end of the sentence (or both), using a single-layer bidirectional GRU network. POS tagging and lemmatization use an averaged perceptron with morphosyntactic features (e.g. word affixes paired with current POS tags) while the dependency parser uses a simple feed-forward neural network to select the optimal transition given by the chosen transition-based dependency parser. Instead of TEPROLIN, UDPipe can be used as a standalone text pre-processing pipeline in RELATE, in production mode (i.e. annotating large corpora using multiple processing threads).

BioNER annotation is obtained through TEPROLIN, by requesting the \texttt{biomedical-named-entity-recognition} operation. This will enable inside-outside-beginning annotations of disorders (\texttt{DISO}), anatomical parts (\texttt{ANAT}), medical procedures (\texttt{PROC}) and drugs and other chemicals (\texttt{CHEM}). \cite{mitrofan2018biomedical} describes the corpus that was used to train a previous version of NLP-Cube to recognize these specialized text spans.

NER annotation is performed using a modified version of an older Stanford Named Entity Recognizer (Stanford NER) \citep{finkel-etal-2005-incorporating} system, making use of CoRoLa pre-trained word embeddings \citep{pais2018corolawe}, computing representations for unknown words based on sub-word information \citep{bojanowski-etal-2017-enriching}. Matching is further enhanced using a rule-based method. Finally, results are attached to a tokenized document using Inside-Outside (IO) notation. This is then converted to Inside-Outside-Beginning (IOB) format for compatibility with other annotations.

Besides TEPROLIN and UDPipe processing, pre-trained language models for other basic language resource kits (BLARK) are available for download from within the RELATE platform\footnote{\url{https://relate.racai.ro/index.php?path=pretrainedlm/list}}. These  models are trained on the same corpus, currently version 2.7 of the RoRefTrees treebank (RRT) \citep{vergirrt2018} available from the Universal Dependencies\footnote{\url{https://universaldependencies.org/}} project. Models are offered for Stanza\footnote{\url{https://stanfordnlp.github.io/stanza/index.html}} \citep{qi-etal-2020-stanza}, RNNTagger\footnote{\url{https://www.cis.uni-muenchen.de/~schmid/tools/RNNTagger/}} \citep{10.1145/3322905.3322915}, NLP-Cube\footnote{\url{https://github.com/adobe/NLP-Cube}} \citep{boros-etal-2018-nlp}, UDPipe \citep{udpipe2016straka} and TreeTagger\footnote{\url{https://www.cis.uni-muenchen.de/~schmid/tools/TreeTagger/}} \citep{Schmidt1994ProbabilisticPT}.

\subsection{Terminology annotation}
\label{sec:terminology}

The terminology annotation tool was initially implemented for the purposes of the international project “Multilingual Resources for CEF.AT in the legal domain” (MARCELL) \footnote{\url{ https://marcell-project.eu/}}. In this context a large comparable corpus \citep{varadi-etal-2020-marcell} of national legislation from the seven partners' countries, including Romania, was enriched with EuroVoc and IATE terminology annotations. Furthermore, a classification of the documents considering the top-level domains of the EuroVoc multilingual thesaurus was done. 

For classifying documents belonging to the Romanian sub-corpus \citep{tufis-etal-2020-collection} we developed a classifier based on the method proposed by \cite{joulin-etal-2017-bag}, using the average of word embeddings over n-grams as features to a linear classifier. The actual implementation used the FastText\footnote{\url{https://fasttext.cc/}} tool. We trained several models, using different word embeddings (CoRoLa, Wikipedia and CommonCrawl). We wanted to compare these models with the well known JEX tool\footnote{\url{https://ec.europa.eu/jrc/en/language-technologies/jrc-eurovoc-indexer}}, developed by the European Commission's (EC) Joint Research Centre (JRC). JEX \citep{steinberger-etal-2012-jrc} is making use of the classification algorithm described in \citep{Pouliquen2006AutomaticAO}, based on a list of lemma frequencies from normalized text, and their weights, that are statistically related to each descriptor, entitled in the paper as associates or as topic signatures. For Romanian, JEX was trained on over 25,000 documents from two parallel corpora: JRC-Acquis \citep{steinberger-etal-2006-jrc} and OPOCE (Publications Office of the European Union). By considering only the first 6 most probable descriptors, it obtained a precision of 45.55\%, a recall of 50.43\% and a F1 score of 47.84\%.

For training and evaluating our EuroVoc models, we used the same approach followed by JEX, employing the same datasets and cross-validation splits. Our best model was obtained using CoRoLa word embeddings and achieved a precision of 50.93\%, a recall of 56.41\% and a F1 score of 53.53\%, thus presenting an increase of over 5\% with regard to F1 score compared to JEX. We further evaluated the model on top level domains and obtained an F1 score of 70.80\%. The model is available for download and interrogation from  RELATE\footnote{\url{https://relate.racai.ro/index.php?path=eurovoc/classify}}. In order to integrate this model in the platform, we employed a modified version of the FastText application, including an embedded web server, allowing the model to be kept in-memory. This server is open sourced\footnote{\url{https://github.com/racai-ai/ServerFastText}}. Besides document classification, term annotation was realized using a NER-based approach, described in \citep{04.11.2019_Tufis_01}.

For the named entities belonging to the location class (LOC) of the  LegalNERo corpus \footnote{\url{https://doi.org/10.5281/zenodo.4772094}} \citep{legalnero_2021}, an enhancement was done linking the named entities with GeoNames database \footnote{\url{https://www.geonames.org/}} by using the feature identifiers associated with each GeoNames feature. GeoNames is a geographical database which contains over 25 million geographical names covering all countries and consists of over 11 million unique placenames, 4.8 million populated places and 13 million alternate names. GeoNames is integrating geographical data such as names of places in various languages, elevation, population and others from various sources. It is available for download. 
For linking the named entities with the GeoNames database a new column was created in the conllup corpus files structure. This GeoNames column was placed last in the file and by default is filled with the “\_” character. This column would hold a value only when the named entity tag is of LOC type and when a perfect match is found within the GeoNames database. The actual match is done via a lookup algorithm implemented in the Python language. In this algorithm, the GeoNames database extract for the Romanian language is loaded into a key:value list, where the key is the standard location name and the value is the GeoNames location code (e.g. for the Suceava city this code is 486885). In many situations one location can have several variations to the naming (that may come from using different spelling or even different names as they are known within different communities). In the GeoNames framework this situation is addressed by having only one standard name (usually the official name of the location) and several secondary names that are part of the GeoNames entry for that location. The script therefore, has created for each such variation a different entry in the lookup dictionary pointing to the same location code value. The conllup file format is token centric, while some locations have a multi-word format. Therefore, the actual first step in the lookup was to build the expressions from the tokens (by finding consecutive LOC tokens). The actual lookup was performed per named entity (a multi-token list). When a perfect match was found (meaning partial sub-expression matches were ignored) all the tokens involved in the match were flagged as a GeoNames entity. The corresponding location code was finally filled in the GeoNames column.

\subsection{Automatic speech recognition}
\label{sec:asr}

The RELATE platform contains two Romanian ASR models that can be accessed using the graphical interface: \texttt{RobinASR} and \texttt{RobinASRDev}. The two variants are based on the DeepSpeech2 architecture \citep{amodei2016deep} and have a reduced number of parameters compared with the original model in order to accommodate to the smaller size of their training corpus. Each model is wrapped as a web service and is deployed on a server with a GPU to achieve low response time to the request of the front-end. However, although the two variants are similar in architecture, they serve different purposes. 

The former variant is a general ASR that was trained on a corpus composed of 230 hours of transcribed audio and obtained a 9.91 word error rate (WER) while combined with a language model \citep{heafield2011kenlm} used for textual correction. The latter variant is a specialized ASR that fine-tunes the general model on the technical domain using the ROBIN Technical Acquisition Speech Corpus (ROBINTASC) corpus \citep{pais_vasile_2021_4626540}, improving the performance by 16.4\% on computer sales conversations. This variant also uses a language model that was trained on the general corpus together with the train ROBINTASC transcriptions, oversampled 10 times to increase n-grams frequencies.

\subsection{Text and speech translation}
\label{sec:translation}

\begin{figure}
    \centering
    \includegraphics[width=0.75\textwidth]{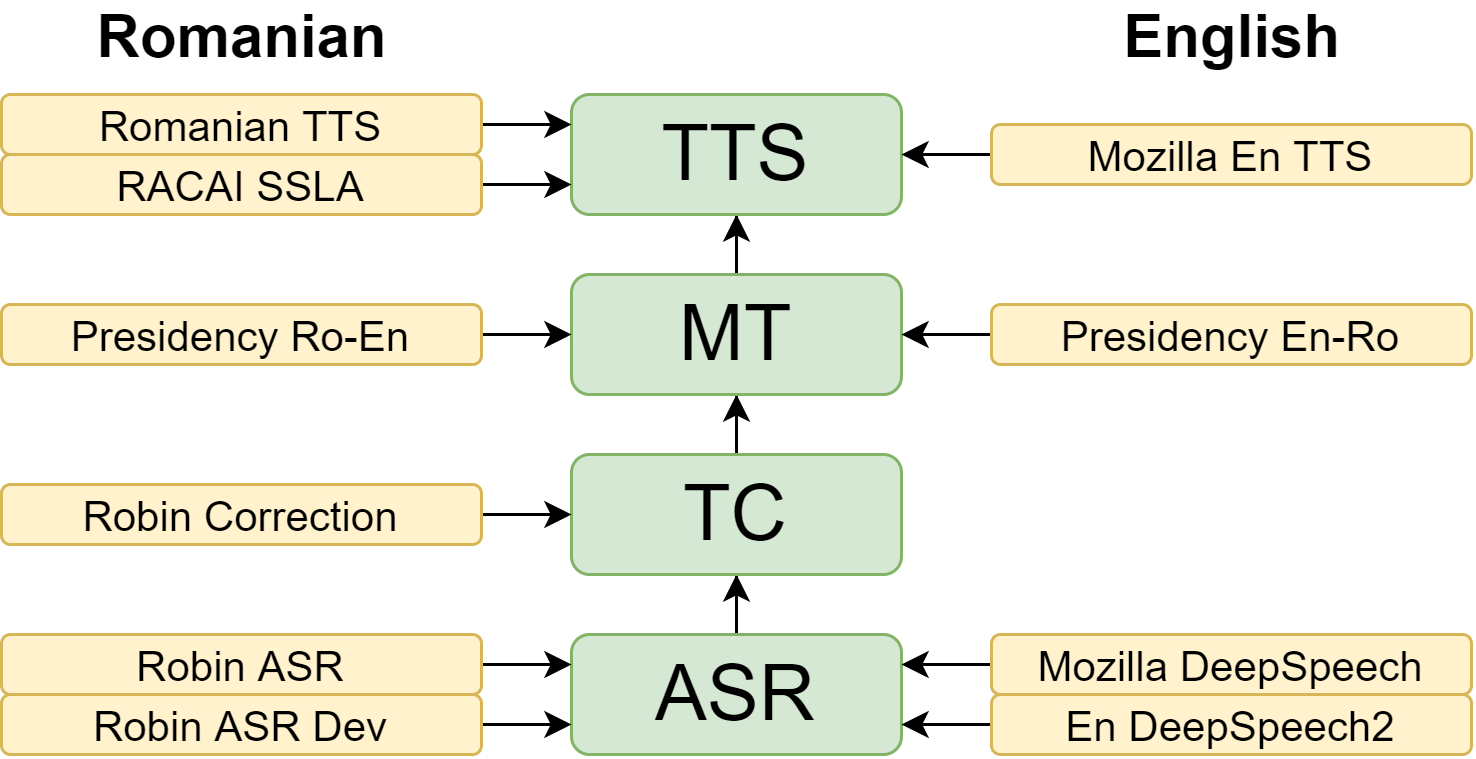}
    \caption{The S2ST cascaded architecture with the four modules and their models (Romanian in the left, English in the right).}
    \label{fig:s2st_arch}
\end{figure}

The speech to speech translation (S2ST) system \cite{avram2021modular} is composed of four cascaded modules: (1) ASR, (2) Textual Correction (TC), (3) Machine Translation (MT) and (4) TTS. Each module contains one or several models that can be configured in a full Romanian-English or English-Romainian inference, as depicted in Figure \ref{fig:s2st_arch}. 

The ASR module contains two models for Romanian - those presented in Subsection \ref{sec:asr} - and two models for English. The first variant - \texttt{En DeepSpeech2} - for English is also based on the DeepSpeech2 architecture and was trained on the LibriSpeech dataset \citep{panayotov2015librispeech}, obtaining a 10.46 WER. The second variant - \texttt{Mozilla DeepSpeech} - for the English is the 0.9.3 version of DeepSpeech \citep{hannun2014deep} offered by Mozilla Speech-To-Text engine\footnote{\url{https://github.com/mozilla/DeepSpeech/releases/tag/v0.9.3}} and that obtained a 7.06\% WER on LibriSpeech.

We offer at this time only a single model in the TC module for Romanian - \texttt{Robin Correction}. Its role is (1) to capitalize the transcribed speech using a list of predefined named entities, (2) replace unknown words with words from a vocabulary and (3) restore hyphen using uni-gram and bi-gram statistics. 

The Romanian-English - \texttt{Presidency Ro-En} - and English-Romanian - \texttt{Presidency En-Ro} - models used by the MT module are the ones developed in the project "CEF Automated Translation toolkit for the Rotating Presidency of the Council of the EU", described in Section \ref{sec:history}.

The TTS module contains a speech synthesizer model for English and two for Romanian. The English version - \texttt{Mozilla En TTS} - is based on Tacotron2 with Dynamic Convolution Attention \citep{battenberg2020location} offered by Mozilla Text-to-Speech\footnote{https://github.com/mozilla/TTS}. Their evaluation outlined a median opinion score (MOS) of 4.31 ± 0.06 in a 95\% confidence interval. The Romanian variants - \texttt{Romanian TTS} and \texttt{RACAI SSLA} - use HMM on their architecture, with the former variant being slower, but producing a higher quality speech and the latter being faster but with a lower quality of the synthesis.

\section{Usage scenarios}
\label{sec:scenarios}
Given the complex nature of the RELATE platform and the large number of integrated components, it has been used for a variety of purposes. However, the scenarios described in this section are the most common ones. Moreover, the description of the usage scenarios provides details into possible interactions between the platform's components.

\subsection{Large corpus processing}
\label{sec:scenario_corpus_processing}
Large corpora can be uploaded in  RELATE as archived .ZIP files. Upon upload, the platform will schedule automatically a task for extracting the archive. Inside the archive, the platform accepts raw text files (with .txt extension) and standoff metadata. Metadata is usually specific to different research projects and can be used by different processing components to embed it into annotated documents.

\begin{figure*}
\includegraphics[width=0.75\textwidth]{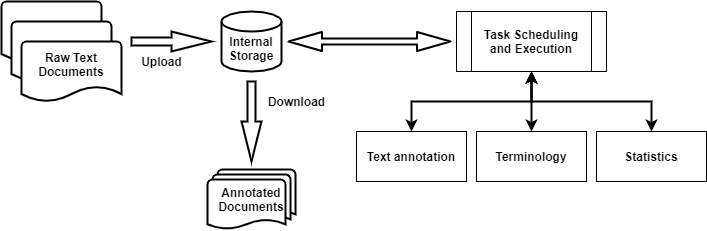}
\caption{Corpus processing flow}
\label{fig:scenario_corpus}
\end{figure*}

\begin{figure*}
\includegraphics[width=0.75\textwidth]{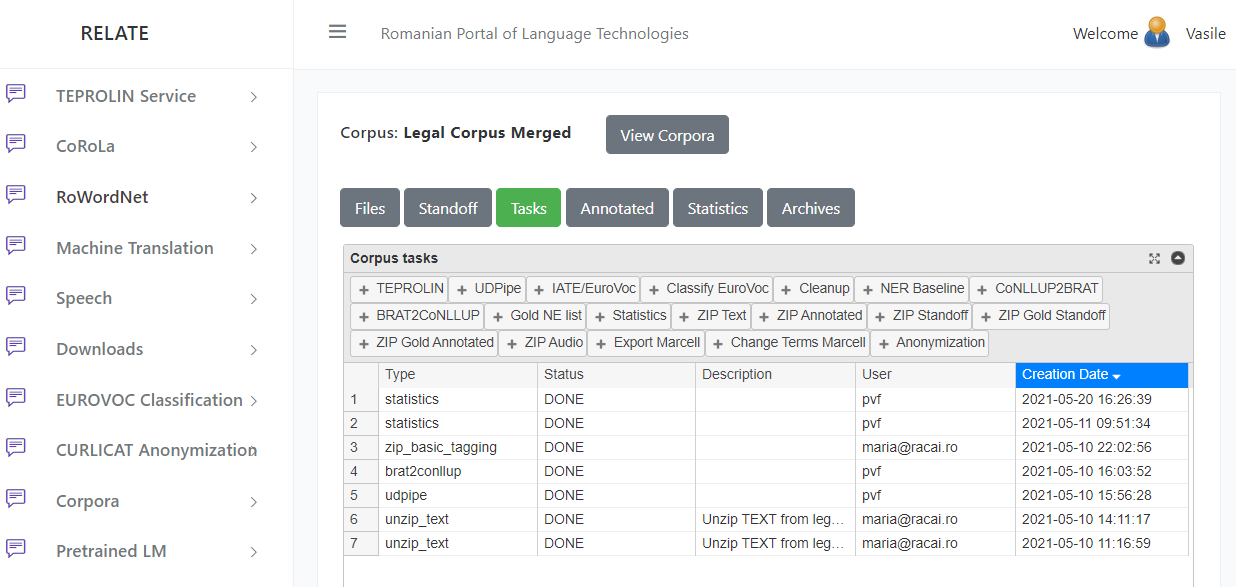}
\caption{Tasks interface component}
\label{fig:scenario_interfata_tasks}
\end{figure*}

Once the documents are available in the platform, different annotation tasks can be scheduled and executed. The tasks interface is presented in Figure \ref{fig:scenario_interfata_tasks}. Furthermore, statistics on both raw text and annotated documents can be computed. The resulting annotated corpus can be archived and downloaded as another .ZIP file. Statistics can be either visualized within the platform or downloaded as .CSV files. The entire flow is depicted in Figure \ref{fig:scenario_corpus}.

\subsection{Creation of human annotated "gold" text corpora}
\label{sec:scenario_manual_corpora}
Creation of "gold" corpora involves human annotators identifying spans of text with certain properties. For the creation of the LegalNERo\footnote{\url{https://doi.org/10.5281/zenodo.4772094}} \citep{legalnero_2021} corpus (a gold corpus for named entity recognition in the Romanian legal domain), we employed five annotators working in RELATE, annotating spans of text with the corresponding named entities (legal reference, person, location, organization and time). 

For the purpose of manual annotation of text corpora,  RELATE  integrates the BRAT\footnote{\url{https://brat.nlplab.org/index.html}} annotation tool \citep{stenetorp-etal-2012-brat}. This allows the user to view one document at a time inside the BRAT component, select with the mouse the desired text span and then associate a corresponding annotation type. The text spans with associated start and end index as well as the annotation type are saved as standoff metadata files.

Since different corpus processing applications may require token based annotations instead of span-based annotations, RELATE  offers dedicated tasks for parallel conversion of the metadata to token-based annotations. This process must be executed after an initial tokenization of the data, possibly including other automatic annotations such as part-of-speech tagging and dependency parsing. The final format for the data will be CoNLL-U Plus files with the gold annotations stored in a dedicated column. For a named entity corpus we use a column named "RELATE:NE". 

\begin{figure*}
\includegraphics[width=0.75\textwidth]{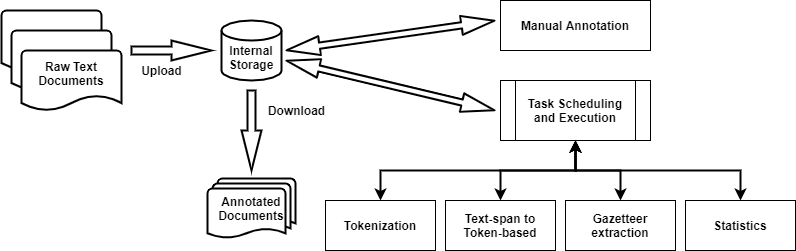}
\caption{Manual corpus annotation processing flow}
\label{fig:scenario_manual_corpora}
\end{figure*}

\begin{figure*}
\includegraphics[width=0.75\textwidth]{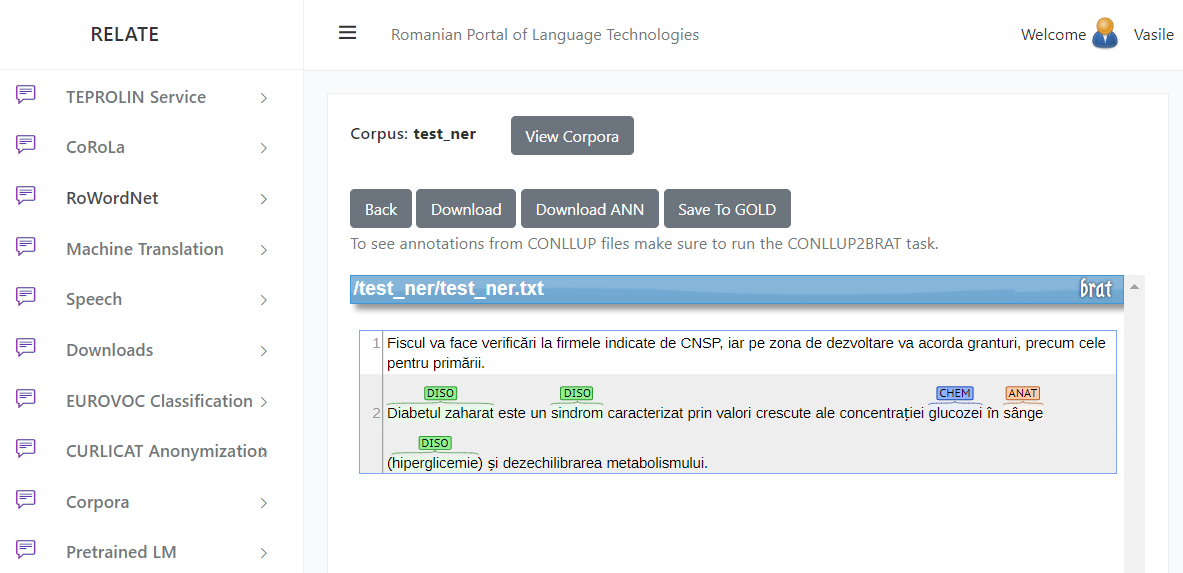}
\caption{Manual corpus annotation GUI}
\label{fig:scenario_interfata_brat}
\end{figure*}

Once the corpus has been annotated (with or without conversion to the token based format), it becomes possible to execute additional tasks within RELATE. A dedicated task allows extraction of gazetteer resources, allowing the produced file to be downloaded and used in other applications requiring such resources. In addition, the statistics component can be used to extract information about the annotated corpus. The processing flow is depicted in Figure \ref{fig:scenario_manual_corpora}, while an example of the manual annotation component used with BioNER entities is presented in Figure \ref{fig:scenario_interfata_brat}.

\subsection{Creation of read speech corpora}
\label{sub:scenario_speech_corpora}
Read speech corpora represent invaluable resources for the creation of Automatic Speech Recognition (ASR) and Text-to-Speech (TTS) systems. These systems can then be used as the building blocks for more advanced processing, like human-machine interaction and speech-to-speech translation. In the context of the ROBIN project, we were concerned with the performance of a low-latency ASR system used for human-robot interaction, considering a micro-world scenario. For this purpose we created the ROBIN Technical Acquisition Speech Corpus (RTASC)\footnote{\url{https://doi.org/10.5281/zenodo.4626539}} \citep{pais_vasile_2021_4626540}, using the audio recording features of the RELATE platform.

\begin{figure*}
\includegraphics[width=0.75\textwidth]{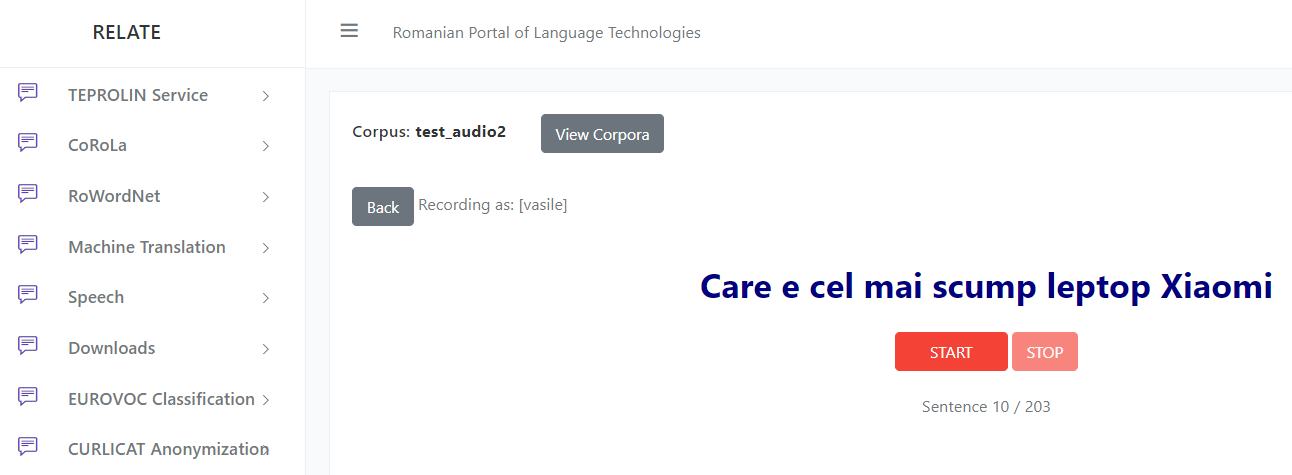}
\caption{The interface used for recording read speech}
\label{fig:scenario_interface_recording}
\end{figure*}

Even though RELATE was initially constructed to process text corpora, the recent addition of speech processing features, allowed it to handle bimodal (text and audio) corpora. For recording purposes, a dedicated component was constructed. It makes use of JavaScript and runs in the user's browser. It displays each sentence from the text component of the corpus and allows the user to record the associated speech. To improve the user experience, the sentence is displayed in a larger font and additional information such as the current sentence number and the total number of sentences is provided. The interface also provides a playback function and, if necessary, allows the user to delete a certain recording in order to record it again. The recording interface is shown in Figure \ref{fig:scenario_interface_recording}.

\begin{figure*}
\includegraphics[width=0.75\textwidth]{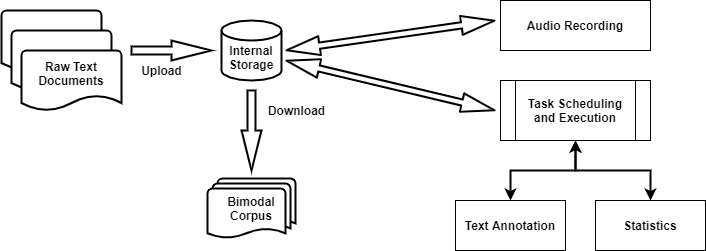}
\caption{Creation of a bimodal corpus}
\label{fig:scenario_audio_recording}
\end{figure*}

Since the text component still plays an important part in a bimodal corpus, all the text processing capabilities can be used to enhance the final corpus. In this case, the annotation pipelines combined with the statistics generation feature are available. The final corpus, comprised of raw text, annotated text and audio files, can be downloaded in an archived format from the platform. The entire processing chain involved is presented in Figure \ref{fig:scenario_audio_recording}.

\section{Conclusions}
\label{sec:conclusions}
In this article we described the development of the RELATE platform, its architecture and considered several of the most common usage scenarios. Currently, the platform provides different levels of processing for Romanian language. Even though initially it was developed for handling large text corpora, recent work added speech capabilities, including speech-to-speech translation for Romanian-English and English-Romanian language pairs.

The platform is designed to be highly customizable and easily extensible by creation of new components. The use of standardized file formats ensures interoperability with other systems. Also, the extensive use of REST APIs allows interoperability with external applications. Finally, the platform is open sourced and available freely on GitHub\footnote{\url{https://github.com/racai-ai/RELATE}}.

The current running version of RELATE\footnote{\url{https://relate.racai.ro}} spans its processing capabilities across four servers hosted at the Institute for Artificial Intelligence "Mihai Dr{\u{a}}g{\u{a}}nescu" of the Romanian Academy. The servers contain both classical CPUs and a GPU resource, allowing different types of language processing tools to be used and guaranteeing high performance.

In the future we are planning to continue integrating new language processing tools in the RELATE platform, as they become available, while aiming to boost interoperability and interactions leading to a productive LT ecosystem \citep{iwltp-2020-international}, at both national and international level. One possible extension in this direction is the replacement of the current EuroVoc annotator with the Romanian model of PyEuroVoc \citep{avram2021pyeurovoc}, a recently introduced legal document classification tool for 22 languages that is based on a more modern architecture - Bidirectional Encoder Representations from Transformers (BERT) \citep{devlin-etal-2019-bert} - and that significantly improved the results of JEX by an average of 28\% F1 score for all languages and 33.8\% for Romanian.




%

%
%

\bibliographystyle{spbasic}      
\typeout{}
\bibliography{bibliography}   

\end{document}